\pdfoutput=1
\documentclass{article} % For LaTeX2e
\usepackage{iclr2027_conference,times}

\usepackage{amsmath,amssymb,amsfonts}
\usepackage{graphicx}
\usepackage{booktabs}
\usepackage{algorithm}
\usepackage{algorithmic}
\usepackage{subcaption}
\usepackage{float}
\usepackage{wrapfig}
\usepackage{xcolor}
\usepackage{enumitem}
\usepackage{makecell}

\usepackage[hidelinks]{hyperref}
\usepackage{url}
\hypersetup{hypertexnames=false}

\newcommand{\vlash}{\textsc{VLASH}}

\newcommand{\pioh}{$\pi_{0.5}$}

\title{\raggedright DEFLECT: Temporal Counterfactual Preference Learning for Delay-Robust Asynchronous VLAs}

\author{
  Yixiang Zhu$^{1,2}$ \quad Yonghao Chen$^{1}$ \quad Jiaxiang Zou$^{1}$ \quad Yusong Hu$^{2}$ \quad Xinyu Chen$^{1}$ \\[4pt]
  $^{1}$The Hong Kong University of Science and Technology (Guangzhou) \\
  $^{2}$One Robotics
}

\iclrfinalcopy

\begin{document}

\maketitle
% Preprint: drop the venue running head that \iclrfinalcopy installs.
\lhead{}
\renewcommand{\headrulewidth}{0pt}

%===============================================================================
\begin{abstract}

Vision-Language-Action (VLA) policies increasingly rely on asynchronous
inference to hide model latency behind ongoing robot motion. During inference,
the robot continues executing its current action chunk while the next chunk is
generated from the observation available at inference start. By the time the
new chunk is executed, the robot or scene may have changed, so an action that
fits the prediction-time state can be misaligned with the execution-time state.
Training-time methods modify the policy input or supervised target based on
the expected execution delay, but do not directly supervise how actions should
change as the scene evolves between prediction and execution.
Runtime repair methods improve chunk continuity or modify actions during
deployment, but can still leave actions aligned with the prediction-time state
rather than the execution-time state.
Our key insight is that execution-time observations are unavailable when the
online prediction is made but are available in recorded trajectories during
offline training, making the prediction--execution mismatch observable during
training. We propose \textsc{DEFLECT}, an offline post-training framework that
turns this temporal contrast into preference supervision for a policy acting
from stale inputs.
Across Kinetix, LIBERO, and three real-robot tasks, \textsc{DEFLECT} improves
delay robustness across the evaluated VLA policies and settings. At the longest
supported delay, it improves success by $4.2$ percentage points over \vlash{}
on Kinetix and by $22.7$ points over naive asynchronous execution with
GR00T N1.7 on LIBERO. 
\end{abstract}

\begin{figure}[h]
\setlength{\abovecaptionskip}{4pt}
\setlength{\intextsep}{6pt}
\centering
\includegraphics[width=0.92\linewidth]{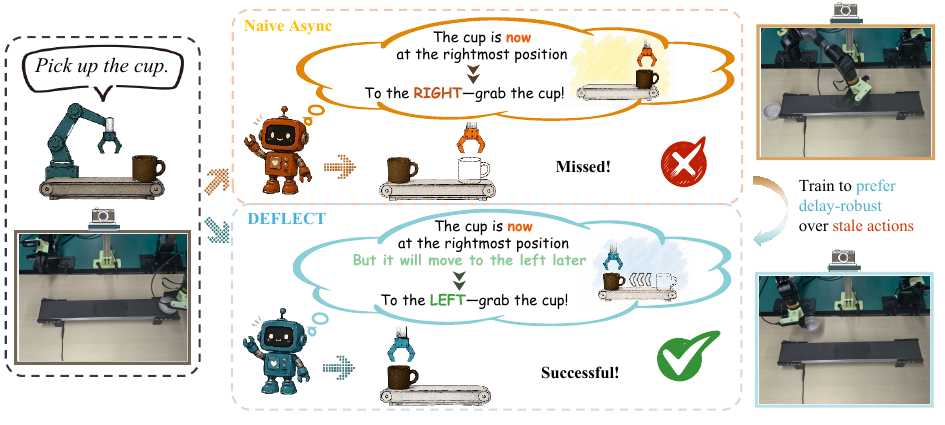}
\caption{\textbf{\textsc{DEFLECT} lets an asynchronous VLA act more robustly from a stale observation.} The task is to grasp a cup on a moving conveyor; between observation and execution the cup travels from right to left. \emph{Top} (Naive Async): the policy reaches where the cup was seen, so the grasp misses. \emph{Bottom} (\textsc{DEFLECT}): given the same stale observation, the policy anticipates the upcoming motion and reaches the new position.}
\label{fig:teaser}
\end{figure}

%===============================================================================

\section{Introduction}
\label{sec:intro}

Vision-Language-Action (VLA) policies have shown strong performance across a wide range of robot manipulation and locomotion tasks~\citep{brohan2023rt, kim2024openvla, black2024pi_0, intelligence2025pi_}. Many of these policies are deployed synchronously and predict actions as \emph{action chunks}~\citep{zhao2023learning}: at each decision step the model outputs a short sequence of actions, the robot executes them in order, and the policy only runs again to produce the next chunk after the current one is finished.
This execution scheme makes robot motion choppy: after each chunk, the robot must wait for the next inference to complete, during which it cannot respond to scene changes.
To remove this pause, \emph{asynchronous inference} has become a common deployment strategy for real-time VLA execution~\citep{tang2025vlash, jiang2511asyncvla, black2026real}, in which the robot keeps executing the current chunk while the model computes the next one in parallel.
Because each action chunk typically takes longer to execute than a single inference pass, the next chunk can be computed before the current one finishes, effectively hiding inference latency behind ongoing robot motion. This overlap also allows the policy to issue new chunks more frequently than stop-and-go execution, improving responsiveness to scene changes.

\begin{figure}[h]
\centering
\includegraphics[width=\linewidth]{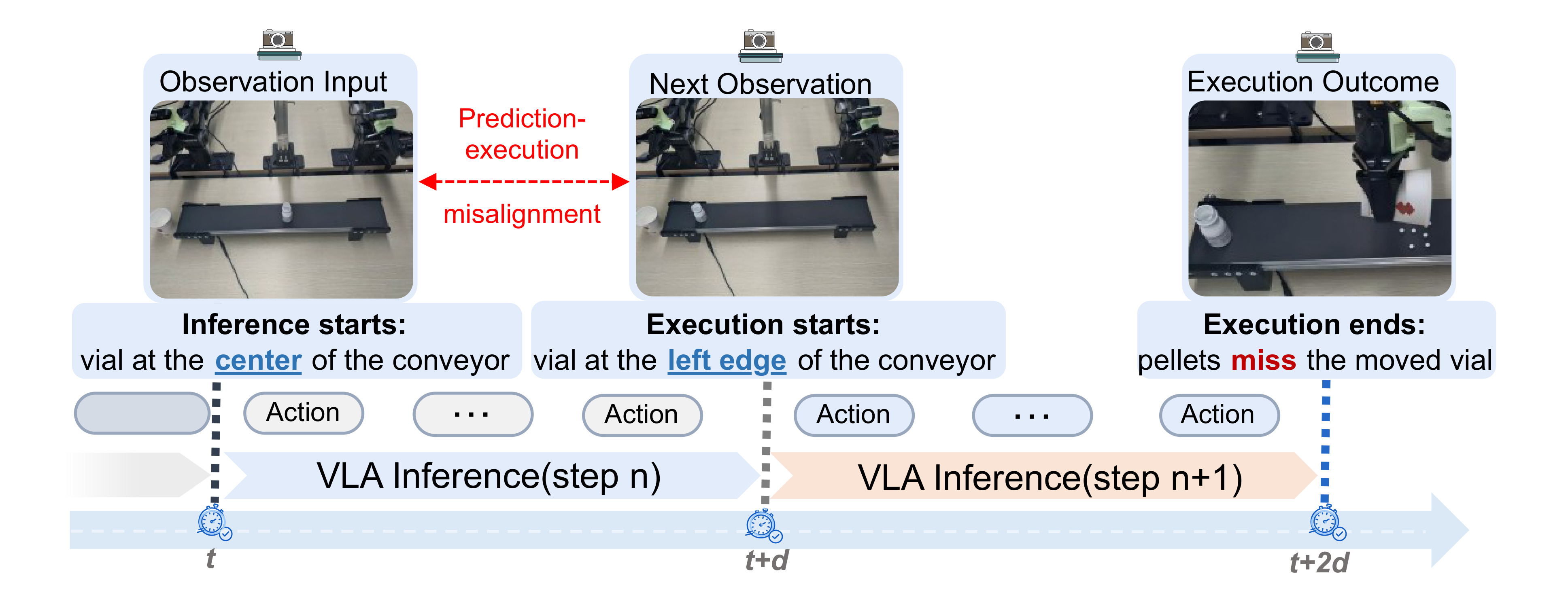}
\caption{\textbf{Async-VLA inference under deployment-relevant latency: a real-robot example.} A robot dispenses pellets into a target vial on a moving conveyor. Top: robot frames at \(t\), \(t+d\), and \(t+2d\). Bottom: consecutive asynchronous VLA inference cycles. The policy starts inference at time \(t\) when the vial is centered, but the resulting actions execute after delay \(d\), when the vial has moved. By execution end, the stale action chunk releases pellets where the vial was, causing a miss.}
\label{fig:motivation}
\end{figure}

However, asynchronous inference also creates a prediction--execution mismatch: the next chunk is computed from an earlier observation but executed after the robot and scene have changed. The resulting action chunk may therefore no longer fit the state the robot actually faces when execution begins. As latency grows, this mismatch accumulates over continuous action execution: each step may look reasonable on its own, but the robot drifts from the correct trajectory and eventually fails the task.
Figure~\ref{fig:motivation} shows the severity of this problem: at prediction time, the vial lies at the center of the conveyor, and the predicted action chunk targets pellet release at that position. By the time the chunk is executed, the vial has moved to the other end of the conveyor, causing the pellets to miss the target. Existing asynchronous strategies~\citep{tang2025vlash, black2026real, liu2025bidirectional, sendai2025leave} can soften this mismatch, but they do not address the central issue: the action chunk should match the execution-time world, not the prediction-time view.

% To address this execution-time alignment gap, we propose \textsc{DEFLECT}, an offline post-training framework that teaches asynchronous VLAs to prefer actions that remain correct when executed. \textsc{DEFLECT} turns prediction--execution mismatch into counterfactual preference supervision: a frozen reference VLA generates a preferred chunk from a future observation, and a rejected chunk from the stale prediction-time observation.
% The trainable policy then scores both chunks under the same stale deployment input used at test time, learning to favor the chunk generated from the future observation over the
% chunk generated from the stale observation.
% This changes the policy's action preference under stale inputs without adding runtime repair, waiting, future-observation access, or extra inference-time computation. 
% To address this mismatch, we propose DEFLECT, an offline post-training framework for delay-robust asynchronous VLAs. A frozen reference VLA generates one chunk from a future observation and another from the stale observation at inference start. We label the future-generated chunk as preferred and train the policy to favor it under the same deployment input used at test time. Future observations are used only to construct the training pairs. The deployment input, asynchronous controller, and inference budget remain unchanged.
We propose DEFLECT, an offline post-training framework for delay-robust asynchronous VLAs. Our key idea is to turn the information advantage of future observations into temporal preference supervision for a policy that must act from stale inputs. A frozen reference VLA generates paired chunks from future and stale observations. The policy learns to prefer the future-generated chunk under its unchanged deployment input. Future observations are used only for pair construction. The controller and inference budget remain unchanged.
Across two simulated benchmarks and three real-robot tasks, \textsc{DEFLECT} improves asynchronous VLA control across the whole delay range its base policy supports, and the margin is largest where latency bites hardest: at the longest such delay it gains $+4.2$ percentage points over \vlash{} on Kinetix and up to $+22.7$ points over naive asynchronous execution on a VLA backbone.

\section{Background and Motivation}
\label{sec:background}

\subsection{Asynchronous VLA Deployment}
\label{sec:async_setup}

We denote the inference delay in control steps by $d$, so a prediction
initiated at timestep $t$ becomes available at $t+d$. At inference start,
the policy receives the visual observation $o_t$, the proprioceptive state
$s_t$, and the language instruction $l_t$. For a policy $b$ obtained through supervised fine-tuning (SFT),
let $\mathcal{C}_b$ denote how it constructs its input during asynchronous
deployment. We use standard SFT for training on the current
observation-state input and the expert action chunk beginning at the same
timestep. Delay-aware SFT instead adjusts the supervised input or
target to account for the expected execution latency.

Let $\mathcal{I}_t$ collect the observation, robot state, instruction, and
any controller-side signals used by $b$. The deployment input $c^{\mathrm{dep}}_{t,b}$ for policy $b$ at control
step $t$ is given by
\begin{equation}
    c^{\mathrm{dep}}_{t,b}
    =
    \mathcal{C}_b(\mathcal{I}_t).
    \label{eq:deployment_context}
\end{equation}
For example, a standard SFT policy deployed with naive asynchronous
execution (Naive Async) uses
$c^{\mathrm{dep}}_{t,b}=(o_t,s_t,l_t)$, while VLASH is a representative
delay-aware SFT method and uses
$c^{\mathrm{dep}}_{t,b}=(o_t,\hat{s}_{t+d},l_t)$, where
$\hat{s}_{t+d}$ is obtained by rolling forward the actions already committed
for execution~\citep{tang2025vlash}. DEFLECT post-trains policy $b$ with the
same deployment input and supervised target. We omit the index $b$ when it
is clear from context.

\subsection{Limitations of Existing Async VLA Strategies}
\label{sec:pitfalls}

% \begin{figure}[t]
%     \centering
%     \includegraphics[width=0.62\linewidth]{figures/framework_a.pdf}
%     \caption{\textbf{Asynchronous deployment contexts under latency.}
%     Inference starts at $s_1$ while the robot continues executing the current action
% chunk $a$, and the next chunk $a'_{0:3}$ is ready at $s_3$: Naive Async uses $(o_1,s_1)$, VLASH/DEFLECT use the deployable $(o_1,s_3)$, and the zero-latency oracle uses the unavailable $(o_3,s_3)$.}
%     \label{fig:async_comparison}
% \end{figure}

\begin{wrapfigure}[20]{r}{0.48\linewidth}
    \vspace{-\baselineskip}
    \centering
    \includegraphics[width=\linewidth]{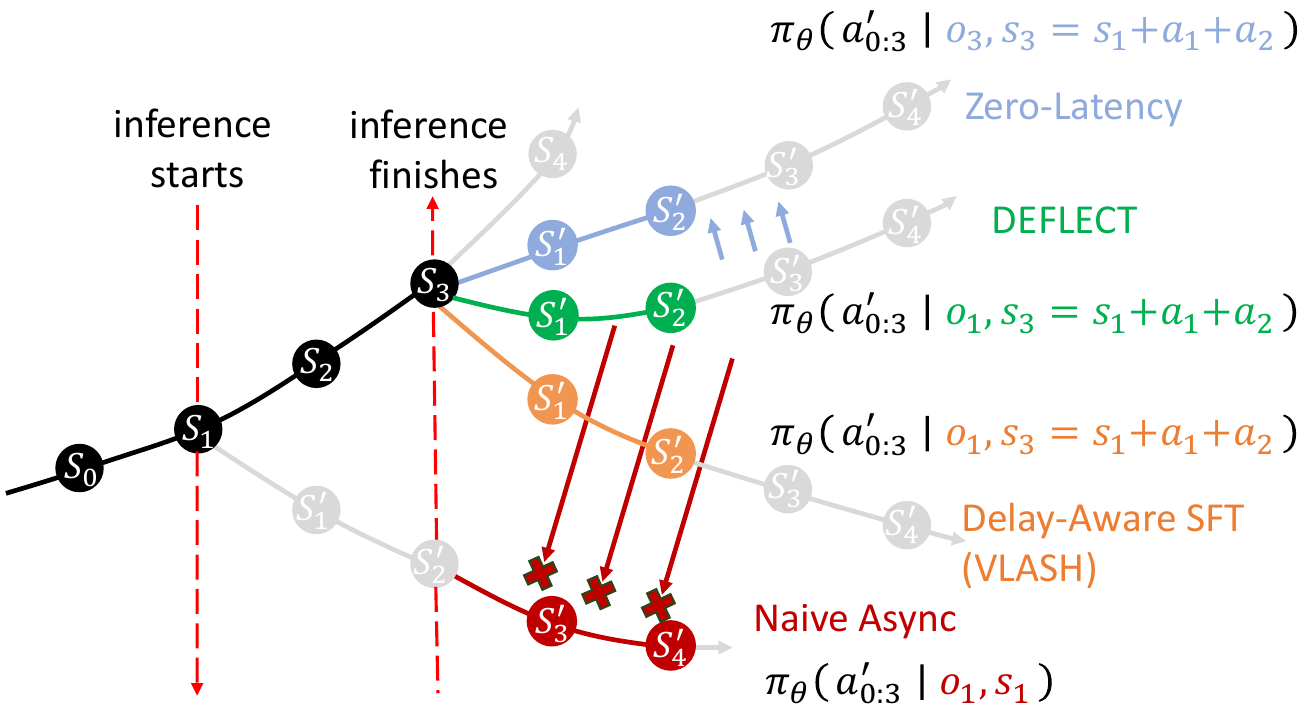}
    \caption{\textbf{Asynchronous deployment contexts under latency.}
Naive Async uses $(o_1,s_1)$, while VLASH represents delay-aware SFT with $(o_1,s_3)$.
The zero-latency oracle uses the unavailable $(o_3,s_3)$.
DEFLECT retains its base SFT policy's deployment input $(o_1, s_3)$.
It learns to favor future-generated over stale-generated chunks,
shifting its actions toward the zero-latency reference.
}
    \label{fig:async_comparison}
    \vspace{-0.5\baselineskip}
\end{wrapfigure}
Existing asynchronous execution strategies mitigate parts of the latency problem, but leave unresolved a more fundamental misalignment: the policy aligns actions with the world observed at prediction time, while the robot needs actions that remain valid at execution time.
As shown in Figure~\ref{fig:async_comparison}, an ideal zero-latency oracle would predict the next chunk $a'_{0:3}$ from the execution-time context $(o_3,s_3)$, but this oracle is unavailable in real deployments: inference for $a'_{0:3}$ starts at $s_1$ while the robot is still executing the current action chunk $a$, and finishes only at $s_3$.
% Naive Async therefore predicts from the fully stale input $(o_1,s_1)$, so its chunk follows the scene at inference start.
% VLASH-style rollout~\citep{tang2025vlash} improves the deployable context to $(o_1,s_3)$, but remains behavior-cloning based, imitating one target rather than ranking execution-time alternatives.
Naive Async therefore predicts from the stale input $(o_1,s_1)$, so its
chunk follows the scene at inference start. Delay-aware SFT methods account
for the expected execution latency during supervised training.
VLASH~\citep{tang2025vlash} is one representative example, using the
rolled-forward deployment context $(o_1,s_3)$ shown in
Figure~\ref{fig:async_comparison}. However, both standard SFT and
delay-aware SFT learn to imitate a single supervised action target for each
input, rather than explicitly ranking execution-time alternatives.
Runtime repair methods such as \textsc{RTC}~\citep{black2026real}, \textsc{BID}~\citep{liu2025bidirectional}, and A2C2~\citep{sendai2025leave} improve chunk continuity.
However, they leave a semantic-timestamp error: an action may be plausible for the prediction-time state yet misaligned with the execution-time state.

\subsection{Our Insight}
\label{sec:proposal}
Our key insight is that future observations are unavailable when an action must
be predicted online, but are available in recorded trajectories during offline
training. They therefore need not be part of the deployment input to influence
the policy's behavior. Instead, later observations can provide information
about which actions are better aligned with the execution-time state and serve
as a basis for learning action preferences, as shown in
Figure~\ref{fig:async_comparison}.

This temporal contrast comes from the observation timestamp. A zero-latency
policy would condition on the execution-time observation
$(o_{t+d},s_{t+d})$, whereas the asynchronous policy must act from the stale
prediction-time view. \textsc{DEFLECT} uses this contrast to define a temporal
counterfactual preference. The chunk generated from the future observation is
labeled as preferred over the chunk generated from the stale view. It then
learns this preference under the same deployment input used at test time,
without changing the runtime interface.
\section{DEFLECT}
\label{sec:deflect_overview}

\begin{figure}[t]
    \centering

    \begin{subfigure}{\linewidth}
        \centering
        \captionsetup{width=0.85\linewidth}
        \includegraphics[width=0.85\linewidth]{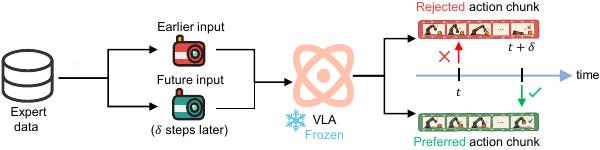}
        \caption{Temporal counterfactual data preparation.
Here $t$ denotes the prediction-time step, and $\delta$ is the sampled
temporal contrast horizon, so the future input corresponds to $t+\delta$.}
        \label{fig:data_prep}
    \end{subfigure}

    \vspace{0.5em}

    \begin{subfigure}{\linewidth}
        \centering
        \captionsetup{width=0.85\linewidth}
        \includegraphics[width=0.85\linewidth]{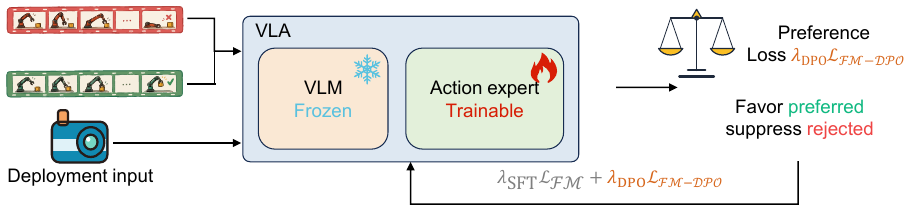}
        \caption{Preference optimization under the deployment input. The objective $\lambda_{\mathrm{SFT}}\mathcal{L}_{\mathrm{FM}}+\lambda_{\mathrm{DPO}}\mathcal{L}_{\mathrm{FM\text{-}DPO}}$ balances the SFT flow-matching anchor and the DPO preference loss.}
        \label{fig:training}
    \end{subfigure}

    \caption{
        \textbf{\textsc{DEFLECT} Overview.}
        (a) A frozen reference VLA contrasts earlier and future inputs from offline trajectories to construct temporal counterfactual supervision.
        (b) \textsc{DEFLECT} freezes the VLM and trains the action expert to favor chunks generated from future observations
over those generated from stale observations, while preserving the expert
action manifold.
    }
    \label{fig:deflect_pipeline}
\end{figure}

Figure~\ref{fig:deflect_pipeline} provides an overview of \textsc{DEFLECT}.
We first construct temporal counterfactual supervision from offline expert trajectories, as shown in Figure~\ref{fig:data_prep}.
The input at $t$ represents the stale prediction-time scene, while
the input at $t+\delta$ represents a future execution-time scene,
where $\delta$ is sampled from the deployment-relevant delay range.
A frozen reference VLA is queried on these two timestamped inputs with shared
sampling randomness, so the resulting contrast mainly reflects the
stale-versus-future input difference rather than sampling noise.

Figure~\ref{fig:training} illustrates how \textsc{DEFLECT} uses this temporal
contrast to train the policy under the deployment-time input
\(c^{\mathrm{dep}}_t\) of its base policy, which is also used at test time.
The stale and future observations are used only to define the offline
preference signal, not as separate optimization inputs.
For the VLA backbones, the VLM backbone is frozen, and only the action expert
is updated with the objective
\[
\lambda_{\mathrm{SFT}}\mathcal{L}_{\mathrm{FM}}
+
\lambda_{\mathrm{DPO}}\mathcal{L}_{\mathrm{FM\text{-}DPO}},
\]
where the Supervised Fine-Tuning (SFT) flow-matching term anchors the policy
to the expert action manifold, and the Flow-Matching Direct Preference Optimization (FM-DPO) term favors the
candidate generated from the future observation over the one generated from
the stale observation.

After training, \textsc{DEFLECT} leaves the asynchronous runtime interface
unchanged. The resulting policy uses the same deployment input and inference
budget as the SFT policy it refines, without access to future observations.

\subsection{Constructing Temporal Counterfactual Preference Pairs} \label{sec:pair_construction} 

DEFLECT constructs temporal preference pairs independently of the deployment
input used by the SFT policy it refines. Let $b$ denote this policy. We use $d$ for the deployment-time inference delay and $\delta$ for the
temporal contrast horizon used to construct a training pair. For each demonstration timestep $t$, we sample a temporal contrast horizon
$\delta$ from the deployment-relevant delay range and query the corresponding
frozen reference policy $\pi_{\mathrm{ref},b}$ using the same language
instruction $l_t$ and sampling randomness $\xi$: \begin{equation} A^+_t = \pi_{\mathrm{ref},b} \big((s_{t+\delta},o_{t+\delta}),l_t;\xi\big), \qquad A^-_t = \pi_{\mathrm{ref},b} \big((s_t,o_t),l_t;\xi\big). \label{eq:preference_pair} \end{equation} $A^+_t$ is generated from the future observation, whereas $A^-_t$ is
generated from the stale observation.
% We label $A^+_t$ as preferred over
% $A^-_t$. The future observation is used only to construct this offline preference and is never provided to the trainable policy. Because both candidates use the same reference policy and sampling randomness, their difference mainly comes from the observation timestamp. 
We label $A_t^+$ as preferred over $A_t^-$. This label is based on the observation timestamps, not on the outcomes of executing the two chunks. It guides training but does not mean that $A_t^+$ performs better in every pair. The future observation is used only to construct the offline preference and is never provided as an input to the trainable policy. Because both candidates use the same reference policy and sampling randomness, their difference mainly comes from the observation timestamp.
The contrast horizon $\delta$ is sampled from the deployment-relevant delay range and independently of any base-policy-specific variable used to construct the deployment input. The pair therefore provides relative temporal supervision rather than a direct regression target for one particular deployment context. 

\subsection{Preference Optimization under the Deployment-Time Input}
\label{sec:deployment_dpo}

\textsc{DEFLECT} retains the deployment input and supervised target of the
SFT policy it refines. Let $\mathcal{D}$ denote the offline demonstration dataset, and let
$\mathcal{S}_b$ denote the sampler used to form the original SFT examples
for policy $b$. For each timestep $t$, it returns an SFT sample
\begin{equation}
    \big(
        c^{\mathrm{dep}}_{t,b},
        A^{\mathrm{exp}}_{t,b}
    \big)
    \sim
    \mathcal{S}_b(\mathcal{D},t),
    \label{eq:base_sft_sample}
\end{equation}
where $c^{\mathrm{dep}}_{t,b}$ is the input used by policy $b$ during
deployment and $A^{\mathrm{exp}}_{t,b}$ is its corresponding expert SFT
target. \textsc{DEFLECT} reuses this input--target pair during post-training
and adds the temporal preference objective.

Although $A^+_t$ and $A^-_t$ are generated from different timestamps,
\textsc{DEFLECT} evaluates both candidates under the same
$c^{\mathrm{dep}}_{t,b}$. This shared context turns the temporal pair into
the decision faced by the deployed policy.

Following DPO and its adaptations to diffusion and flow-matching models
\citep{rafailov2023direct,wallace2024diffusion,mcallister2026flow}, we
estimate the per-example flow-matching objective using one sampled
$(\tau,\epsilon)$ and use the resulting loss as a surrogate score for
negative log likelihood. Let $\theta$ denote the trainable policy parameters and
$v_\theta(\cdot)$ its predicted flow velocity field. For an action chunk $A$, conditioning input $c$,
flow time $\tau\in[0,1]$, and Gaussian noise $\epsilon$, define
\begin{equation}
    A_\tau
    =
    \tau A+(1-\tau)\epsilon,
    \qquad
    \mathcal{L}_{\mathrm{FM}}
    (\theta;A,c,\tau,\epsilon)
    =
    \left\|
        v_\theta(A_\tau,\tau,c)-(A-\epsilon)
    \right\|_2^2 .
    \label{eq:fm_loss}
\end{equation}
The same sampled $(\tau,\epsilon)$ is reused across both candidates, the
trainable and reference policies, and the SFT anchor. Define
\begin{equation}
\begin{aligned}
    \Delta_\theta^\pm
    &=
    \mathcal{L}_{\mathrm{FM}}
    (\theta;
     A_t^\pm,
     c_{t,b}^{\mathrm{dep}},
     \tau,\epsilon)
    -
    \mathcal{L}_{\mathrm{FM}}
    (\pi_{\mathrm{ref},b};
     A_t^\pm,
     c_{t,b}^{\mathrm{dep}},
     \tau,\epsilon),\\
    \mathcal{L}_{\mathrm{FM\text{-}DPO}}(\theta)
    &=
    -\log\sigma
    \left(
        \beta
        \left[
            \Delta_\theta^- - \Delta_\theta^+
        \right]
    \right),
\end{aligned}
\label{eq:dpo_fm}
\end{equation}
where $\sigma$ is the sigmoid function and $\beta$ controls the preference
strength. This objective provides a reference-calibrated flow-matching
surrogate for the DPO log-ratio margin. Under the deployment input of policy $b$, it favors the candidate generated
from the future observation over the candidate generated from the stale
observation.

The final objective combines this relative preference with the original
supervised target:
\begin{equation}
    \mathcal{L}_{\mathrm{DEFLECT}}(\theta)
    =
    \lambda_{\mathrm{SFT}}
    \mathcal{L}_{\mathrm{FM}}
    (\theta;
     A^{\mathrm{exp}}_{t,b},
     c^{\mathrm{dep}}_{t,b},
     \tau,\epsilon)
    +
    \lambda_{\mathrm{DPO}}
    \mathcal{L}_{\mathrm{FM\text{-}DPO}}(\theta).
    \label{eq:deflect}
\end{equation}
Here $\lambda_{\mathrm{SFT}}\geq0$ weights the supervised term and
$\lambda_{\mathrm{DPO}}\geq0$ the preference term. The SFT term preserves the supervised behavior of policy $b$, while the
preference term introduces a future-aware temporal correction. When
$\lambda_{\mathrm{DPO}}=0$, Eq.~\eqref{eq:deflect} reduces to continued SFT
of the same policy. For policies deployed with Naive Async, one checkpoint
serves all evaluated delay settings using the same input form
$(o_t,s_t,l_t)$. This setup yields a correction shared across the training
delay range. At deployment, \textsc{DEFLECT} retains the context constructor,
asynchronous controller, and inference budget of the SFT policy it refines.
\section{Experiments}
\label{sec:experiments}

We design our experiments and analyses to answer the following four questions:
\begin{enumerate}[leftmargin=2.0em]
    \item Does \textsc{DEFLECT} improve task success over naive asynchronous
    execution and strong baselines in simulation and the real world, without
    adding inference-time computation?

    \item Does \textsc{DEFLECT} improve over Continued SFT and Future SFT, and which
    design choices are important for its performance?

    \item Does \textsc{DEFLECT} improve both standard and delay-aware SFT
    policies across different VLA backbones and simulated and real-robot
    settings?

    \item How does \textsc{DEFLECT}'s benefit change with inference delay, and
    how sensitive is it to the contrast-horizon range used to construct
    preference pairs?
\end{enumerate}

\subsection{Simulation Evaluation} \label{sec:simulation} We evaluate DEFLECT on two simulated robot-control benchmarks: Kinetix~\citep{matthews2410kinetix} and LIBERO~\citep{liu2023libero}. Kinetix focuses on fast reactions in highly dynamic environments, while LIBERO evaluates image-based manipulation across a diverse set of tasks. \subsubsection{Kinetix} \label{sec:kinetix} \paragraph{Experimental Setup.} Kinetix~\citep{matthews2410kinetix} contains 12 dynamic control tasks involving contact-rich manipulation and locomotion, including catching, launching, and balancing. Since the robot and the environment can change substantially during a single inference call, Kinetix provides a direct test of robustness to prediction--execution delay. Following RTC~\citep{black2026real} and VLASH~\citep{tang2025vlash}, we use an action-chunking flow policy with an action-chunk length of
$H=8$ and a four-layer MLP-Mixer architecture. All methods in the standard comparison use the same released VLASH checkpoint, action-chunk length, and asynchronous execution protocol. Following the VLASH evaluation convention, we use an execution horizon of
$K=\max(1,d)$, meaning that $K$ actions are executed from each chunk before
switching to the next prediction. We evaluate simulated inference delays
$d\in\{0,1,2,3,4\}$. We report the average success rate across all 12 tasks, with 1,024 rollouts per task for each delay and method. \textsc{DEFLECT} is post-trained for 24 epochs.

\paragraph{Baselines.} We compare DEFLECT against the following baselines and controls: 
\begin{itemize}[leftmargin=2.0em]

\item \textbf{Naive.} This baseline runs policy inference in parallel with action execution and switches to the newly predicted chunk as soon as it becomes available. It does not explicitly correct for the stale observation used to predict the new chunk. 

\item \textbf{RTC.} Real-Time Chunking~\citep{black2026real} keeps the actions that are already committed for execution and inpaints the remaining part of the new chunk. This repairs chunk transitions but requires additional computation during deployment. 

\item \textbf{BID.} Bidirectional Decoding~\citep{liu2025bidirectional} resamples action chunks at test time using forward-contrast and backward-coherence criteria. Like RTC, it modifies the runtime sampling procedure and adds inference-time computation. 

\item \textbf{VLASH.} VLASH~\citep{tang2025vlash} is a training-time asynchronous baseline that rolls the proprioceptive state forward to the expected execution time while retaining the visual observation captured at inference start. For this benchmark, its released checkpoint serves as the delay-aware SFT
initialization for \textsc{DEFLECT}.

\end{itemize}

We additionally compare with TT-RTC~\citep{black2025trainingtimeactionconditioningefficient} and REMAC~\citep{wang2026real} in a separate protocol-matched experiment. REMAC reports results by averaging over all valid execution horizons, whereas the standard VLASH protocol evaluates the single horizon $K=\max(1,d)$. We therefore evaluate all methods under REMAC's convention for this comparison rather than mixing results from the two protocols.

\begin{figure}[t]
\centering
\includegraphics[width=\linewidth]{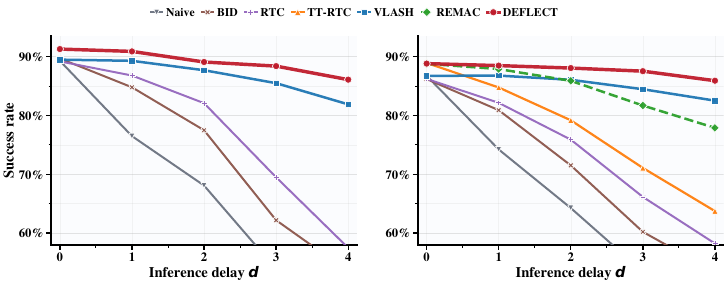}\\[-0.55em]
\begin{minipage}[t]{0.53\linewidth}
    \centering
    \subcaption{Standard protocol ($K{=}\max(1,d)$)}
    \label{fig:kinetix_protocols_a}
\end{minipage}%
\begin{minipage}[t]{0.47\linewidth}
    \centering
    \subcaption{\textsc{REMAC} protocol ($K$ averaged)}
    \label{fig:kinetix_protocols_b}
\end{minipage}
\vspace{-0.9em}
\caption{\textbf{Kinetix success rate} under the two evaluation conventions. Averaging over
all legal horizons $K \in [\max(1,d),\,8-d]$ in (b) lowers every method's absolute values
relative to (a). }
\label{fig:kinetix_protocols}
\end{figure}

\paragraph{Results.}
As shown in Figure~\ref{fig:kinetix_protocols}(a), DEFLECT achieves the best performance
under the VLASH protocol, improving the average success rate from $86.8\%$ to
$89.2\%$ over VLASH. At $d=4$, DEFLECT reaches $86.1\%$, outperforming
VLASH by $4.2$ percentage points. Figure~\ref{fig:kinetix_protocols}(b) shows that
the conclusion remains unchanged under the REMAC protocol: DEFLECT achieves
the highest average success rate of $87.8\%$ and exceeds REMAC by $8.0$
points at $d=4$. Ablations show that keeping the SFT loss and scoring both candidates
with the same deployment input improve task success (Appendix~\ref{app:ablations}).
Even when preference pairs use observations only one or two control
steps apart, \textsc{DEFLECT} still outperforms \textsc{VLASH} by $3.2$ percentage points
at a four-step inference delay (Appendix~\ref{app:zeroshot}).

\subsubsection{LIBERO}
\label{sec:libero}

\paragraph{Experimental Setup.} We evaluate DEFLECT on LIBERO~\citep{liu2023libero}, a widely used VLA benchmark containing four task suites: Spatial, Object, Goal, and LIBERO-10, with ten tasks in each suite. We evaluate three SFT-trained VLA policies: $\pi_{0.5}$, SmolVLA, and
GR00T N1.7. The $\pi_{0.5}$ policy uses a delay-aware SFT initialization
following VLASH, while SmolVLA and GR00T N1.7 retain their standard SFT
initializations and use Naive Async at deployment. \textsc{DEFLECT} preserves
the deployment input and supervised target of each policy. We follow the
standard evaluation protocol of each backbone, using execution horizons of $K=5$ for $\pi_{0.5}$ and SmolVLA, and
$K=8$ for GR00T N1.7. We evaluate
delays $d\in\{1,2,3,4\}$, using 500 episodes per suite and delay for all
three backbones.

\begin{table*}[t]
\centering
\caption{
\textbf{LIBERO success rates (\%) across VLA policies.}
Avg.\ denotes the mean over delays $d\in\{1,2,3,4\}$, measured in control steps.
In (b), the base policy is VLASH for $\pi_{0.5}$ and Naive Async for SmolVLA and GR00T N1.7.
Gains are relative to the base policy and reported in percentage points (pp).
}
\label{tab:libero_main}
\small
\setlength{\tabcolsep}{6pt}

\textbf{(a) GR00T N1.7: Delay sweep and SFT baselines}

\vspace{0.3em}

\begin{tabular*}{\linewidth}{@{\extracolsep{\fill}}lccccc@{}}
\toprule
Method & $d=1$ & $d=2$ & $d=3$ & $d=4$ & Avg. \\
\midrule
Naive Async
    & $92.25$ & $84.90$ & $69.25$ & $50.40$ & $74.20$ \\
Continued SFT ($\lambda_{\mathrm{DPO}}=0$)
    & $91.65$ & $85.40$ & $75.55$ & $62.25$ & $78.71$ \\
Future SFT
    & $\mathbf{95.20}$ & $90.10$ & $80.60$ & $65.75$ & $82.91$ \\
\addlinespace[2pt]
\textsc{DEFLECT}
    & $93.60$ & $\mathbf{91.05}$ & $\mathbf{85.30}$ & $\mathbf{73.10}$ & $\mathbf{85.76}$ \\
\bottomrule
\end{tabular*}

\vspace{0.8em}

\textbf{(b) Gains across VLA backbones and SFT initializations}

\vspace{0.3em}

\begin{tabular*}{\linewidth}{@{\extracolsep{\fill}}llcccc@{}}
\toprule
 & & \multicolumn{2}{c}{Avg. success rate (\%)} & \multicolumn{2}{c}{Gain (pp)} \\
\cmidrule(lr){3-4}\cmidrule(lr){5-6}
Backbone & SFT initialization & Base policy & \textsc{DEFLECT} & Avg. & $d=4$ \\
\midrule
$\pi_{0.5}$ & Delay-aware (VLASH)
    & $95.54$ & $96.28$ & $+0.74$ & $+1.40$ \\
SmolVLA & Standard
    & $49.77$ & $60.26$ & $+10.49$ & $+14.00$ \\
GR00T N1.7 & Standard
    & $74.20$ & $85.76$ & $+11.56$ & $+22.70$ \\
\bottomrule
\end{tabular*}
\end{table*}

\paragraph{Results.}
As shown in Table~\ref{tab:libero_main}(a), DEFLECT substantially improves
the delay robustness of GR00T N1.7. Its gains grow with the delay: $+1.35$ percentage points
at $d=1$, $+6.15$ at $d=2$, $+16.05$ at $d=3$, and $+22.70$ at $d=4$. As a result,
the average success rate increases from $74.20\%$ to $85.76\%$.
% Continued SFT reaches $78.71\%$, while DEFLECT provides a further
% $+7.05$ percentage points on average. The additional benefit over continued
% SFT is concentrated at $d\geq2$, where the prediction and execution states
% differ more substantially.

Future SFT performs best at $d=1$, but its success decreases more sharply as delay grows. DEFLECT exceeds it by 2.85 percentage points on average and 7.35 points at $d=4$, suggesting that preference learning improves delay robustness beyond directly learning future-generated targets.

Table~\ref{tab:libero_main}(b) further shows positive gains across all three
SFT policies.
\textsc{DEFLECT} improves the delay-aware $\pi_{0.5}$ policy by $+0.74$
percentage points on average, standard-SFT SmolVLA by $+10.49$, and
standard-SFT GR00T N1.7 by $+11.56$. Because the backbones and
evaluation protocols differ, we evaluate the improvement over each policy's
own SFT baseline rather than directly comparing gain magnitudes across
backbones. The consistent improvements show
that \textsc{DEFLECT} can refine both standard and delay-aware SFT policies
across VLA backbones.

\subsection{Real-World Evaluation}
\label{sec:real_world}

To evaluate \textsc{DEFLECT} in real-world settings, we deploy an in-house
$\pi_{0.5}$ policy on a bimanual AgileX PiperX platform with two 6-DoF arms.
The robot is equipped with three Intel RealSense D435i cameras, including one
overhead camera and one wrist camera on each arm. Our system is built on
LeRobot~\citep{cadene2026lerobot}. The comparison covers three stages of the
same $\pi_{0.5}$ training pipeline: standard SFT with Naive Async,
delay-aware SFT following VLASH, and \textsc{DEFLECT} post-training initialized
from the delay-aware SFT policy. We evaluate task success in dynamic
multi-stage manipulation and responsiveness to rapid environmental changes.

\subsubsection{Task Success and Reactivity}
\label{sec:real_world_tasks}

\paragraph{Experimental Setup.}
We evaluate $\pi_{0.5}$ with an action-chunk length of $H=50$ and
10 denoising steps. The robot operates at 30 Hz, and the asynchronous
controller requests a new prediction when half of the current action queue
remains.  We compare
$\pi_{0.5}$ with Naive Async execution, VLASH, and DEFLECT. All methods
use the same demonstrations, VLA backbone, asynchronous controller, and
inference budget. 

We evaluate the following three tasks:

\begin{itemize}[leftmargin=2.0em]
    \item \textbf{Conveyor-I.}
    The left arm places a cup on a moving conveyor, and the right arm
    retrieves it before it leaves the workspace.

    \item \textbf{Conveyor-II.}
    The left arm places a cup on the conveyor. The right arm then selects
    the red cube rather than the green cube, places it into the moving cup,
    and retrieves the cup.

    \item \textbf{Whack-a-Mole.}
    The robot strikes targets that appear briefly at random positions before
    they retract.
\end{itemize}

For each task, we conduct 30 trials per method. We report full-task success
rates for Conveyor-I and Conveyor-II, where every sub-goal must be completed,
and the average number of targets struck during each 30-second
Whack-a-Mole trial. Detailed training configurations and per-sub-goal results
are provided in the appendix.

\begin{figure}[t]
\centering
\includegraphics[width=0.96\linewidth]{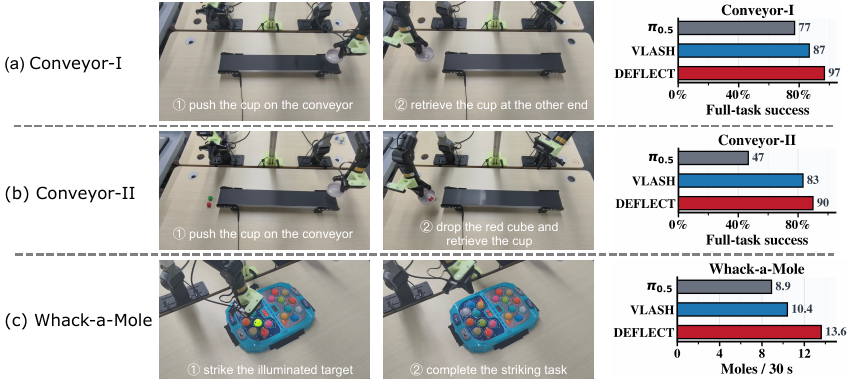}
\caption{
\textbf{Real-robot tasks and results.}
(a) \emph{Conveyor-I}, (b) \emph{Conveyor-II}, and
(c) \emph{Whack-a-Mole}.
The conveyor tasks report full-task success, while Whack-a-Mole reports
the average number of targets struck per 30-second trial.
}
\label{fig:realrobot}
\end{figure}

\paragraph{Results.}
As shown in Figure~\ref{fig:realrobot}, DEFLECT outperforms both
$\pi_{0.5}$ with Naive Async and VLASH across all three tasks. On
Conveyor-I, DEFLECT achieves a full-task success rate of $96.7\%$, compared
with $86.7\%$ for VLASH and $76.7\%$ for $\pi_{0.5}$. On Conveyor-II,
DEFLECT reaches $90.0\%$, compared with $83.3\%$ for VLASH and $46.7\%$
for $\pi_{0.5}$. On Whack-a-Mole, DEFLECT strikes an average of $13.6$ targets
per 30-second trial, outperforming VLASH at $10.4$ and $\pi_{0.5}$ at
$8.9$. These results show that DEFLECT improves both multi-stage
manipulation and reactive control without additional inference-time
computation.

% \begin{table}[t]
% \centering
% \small
% \caption{
% \textbf{Inference-latency breakdown for \pioh{}.}
% Times are in milliseconds. Delays are rounded up to whole control steps at $30$\,Hz.
% }
% \label{tab:latency}
% \begin{tabular*}{0.9\linewidth}{@{\extracolsep{\fill}}lrrr@{}}
% \toprule
%  & \textbf{RTX 5090} & \textbf{RTX 4090} & \textbf{RTX 5070} \\
% \midrule
% Vision encoder + VLM prefill
%     & $38.3$ & $56.9$ & $116.6$ \\
% Action expert (10 steps)
%     & $18.8$ & $21.9$ & $35.4$ \\
% \midrule
% % \addlinespace[2pt]
% \textbf{Model inference}
%     & $\mathbf{57.1}$ & $\mathbf{78.8}$ & $\mathbf{152.0}$ \\
% \quad Model delay (steps)
%     & $2$ & $3$ & $5$ \\
% \bottomrule
% \end{tabular*}
% \end{table}

\begin{table}[t]
\centering
\small
\caption{
\textbf{Model inference latency of \pioh{} across GPUs.}
The action expert uses 10 denoising steps. Equivalent delay is the total
inference latency rounded up to whole control steps at $30$\,Hz.
}
\label{tab:latency}
\begin{tabular*}{\linewidth}{@{\extracolsep{\fill}}lcccc@{}}
\toprule
 & \multicolumn{3}{c}{Inference latency (ms)} & \\
\cmidrule(lr){2-4}
GPU & \makecell{Vision encoder\\+ VLM prefill} & Action expert & \textbf{Total}
    & \makecell{Equivalent delay\\(control steps)} \\
\midrule
RTX 5090 & $38.3$  & $18.8$ & $\mathbf{57.1}$  & $2$ \\
RTX 4090 & $56.9$  & $21.9$ & $\mathbf{78.8}$  & $3$ \\
RTX 5070 & $116.6$ & $35.4$ & $\mathbf{152.0}$ & $5$ \\
\bottomrule
\end{tabular*}
\end{table}

\subsubsection{Model Inference Latency}
% \subsubsection{Deployment Latency}
\label{sec:deployment_latency}

Table~\ref{tab:latency} reports the inference latency of \pioh{} across three GPUs. 
On the RTX 5090 used in our real-robot experiments, model inference takes $57.1$\,ms; it takes $78.8$\,ms on the RTX 4090 and $152.0$\,ms on the RTX 5070. 
Rounding these latencies up to whole control periods at $30$\,Hz gives two steps for the RTX 5090, three for the RTX 4090, and five for the RTX 5070. 
These hardware latencies are comparable in scale to the multi-step delays evaluated in simulation.
FASTER~\citep{lu2026faster} reduces the \pioh{} time-to-first-action to $62.1$\,ms on an RTX 4090, still nearly two control periods at $30$\,Hz. 
Such residual latency motivates delay-robust control alongside faster inference, which is the focus of \textsc{DEFLECT}.

% \yh{Table~\ref{tab:latency} reports the model inference latency of \pioh{} across three GPUs.
% On the RTX 5090 used in our real-robot experiments, inference takes $57.1$\,ms, corresponding to two control steps after rounding up. 
% The equivalent delays are three and five steps on the RTX 4090 and RTX 5070, respectively. 
% Model inference alone thus spans multiple control periods, supporting the relevance of the multi-step delays evaluated in simulation. 
% FASTER~\citep{lu2026faster} reports a time-to-first-action of $62.1$\,ms for \pioh{} on an RTX 4090, still nearly two control periods at $30$\,Hz. 
% This residual latency motivates \textsc{DEFLECT}'s focus on delay-robust control as a complement to inference acceleration.}
\section{Conclusion}
\label{sec:conclusion}

% We present \textsc{DEFLECT}, an offline post-training framework for
% delay-robust asynchronous Vision-Language-Action models with
% flow-matching action heads. By contrasting action chunks generated
% from future and stale observations under the same deployment input,
% \textsc{DEFLECT} learns to better handle the prediction--execution
% mismatch. Experiments on Kinetix and LIBERO demonstrate improved
% asynchronous control across the evaluated policies, with the largest
% gains at longer delays and no additional inference-time computation.
% Real-robot experiments further show improvements in moving-object
% manipulation and reactive control. We hope these results encourage
% further exploration of temporal preference learning for more reliable
% VLA control in dynamic environments.
We present \textsc{DEFLECT}, an offline post-training framework for
delay-robust asynchronous Vision-Language-Action models with
flow-matching action heads. By contrasting action chunks generated
from future and stale observations under the same deployment input,
\textsc{DEFLECT} learns to reduce the prediction--execution mismatch.
Across Kinetix, LIBERO, and real-robot tasks, \textsc{DEFLECT} improves
asynchronous control, with larger gains at longer delays and no
additional inference-time computation. These results support temporal
preference learning for more reliable VLA control in dynamic environments.

%===============================================================================
% ICLR statements (do not count toward the page limit)

% =============================================================================
% ICLR 2027 required / recommended statements.
% These do NOT count toward the 9-page main-text limit.
% TODO(authors): verify that the AI use statement below matches your actual
% workflow before submission. The ICLR AI Policy for Authors requires this
% section and holds the authors responsible for its accuracy.
% =============================================================================

\subsection*{Reproducibility statement}

All components needed to reproduce our results are specified in the paper.
The \textsc{DEFLECT} objective is given in closed form in Eqs.~\eqref{eq:preference_pair}--\eqref{eq:deflect}, and one full gradient step, including preference-pair construction, the shared $(\tau,\epsilon)$ sampling protocol, and the joint SFT$+$DPO update, is written out as pseudocode in Algorithm~\ref{alg:deflect_step} (Appendix~\ref{app:algorithm}).
Appendix~\ref{app:implementation} lists the reference checkpoints, trainable-parameter counts, optimizer, schedule, batch size, learning rate, and both loss weights for each benchmark (Table~\ref{tab:hyperparams}), together with the flow-matching loss-reduction convention that the DPO margin depends on.
Evaluation protocols and episode counts are given in Appendix~\ref{app:implementation}.
Both benchmarks (Kinetix~\citep{matthews2410kinetix} and LIBERO~\citep{liu2023libero}) are public, and we use their standard reward, termination, episode-length, and success conventions without modification.
The real-robot experiments depend on our specific bimanual hardware setup, described in Appendix~\ref{app:realrobot}; those results are reproducible only on comparable hardware.

\subsection*{AI use statement}

In this work, we used generative AI tools in three roles, and we describe all of them in full.

For writing, we used them for grammar correction, phrasing, and copy-editing of author-written text.

For code, we used an AI coding assistant throughout the experimental pipeline rather than for plotting alone. The preference-pair collection, post-training, and evaluation scripts, the cluster job-submission and monitoring tooling, and the analysis and figure-plotting code were written with AI assistance. The assistant was also used to launch these jobs, to parse the resulting run logs, and to compute the summary statistics and significance tests we report. In the course of that work it also proposed diagnostic experiments and ablations, several of which we ran and several of which we rejected.

We also used generative AI tools to search for and summarize related literature. Every work cited in this paper was then located and read by at least one author, and we did not rely on a model's summary for any claim we attribute to prior work.
The research question and the \textsc{DEFLECT} objective are the authors' own.
All AI-assisted text was reviewed and edited by the authors. All AI-assisted code was read by at least one author, and every number reported in this paper is the output of executing that code on the public benchmarks or on our own robot, checked against the raw run logs rather than transcribed from a model's summary.
We take responsibility for the final content of this work, including all text, claims, and artifacts.

%===============================================================================

\bibliography{references}
\bibliographystyle{iclr2027_conference}

%===============================================================================
% Appendix
\clearpage
\appendix

\section{Related Work}
\label{sec:related}

\paragraph{Efficient VLA inference.} Recent VLAs commonly predict action chunks using diffusion or flow matching \citep{zhao2023learning,black2024pi_0,intelligence2025pi_}. To improve real-time deployment, DeeR-VLA reduces backbone computation, SnapFlow reduces flow sampling steps, and FASTER prioritizes and streams near-term actions \citep{yue2024deer,luan2026snapflow,lu2026faster}. These methods shorten the time before actions become available, but do not eliminate the prediction--execution mismatch. Nonzero latency remains, and later actions in a chunk are still conditioned on an earlier observation, so they may become stale before execution. DEFLECT instead learns actions that remain aligned with the scene after this residual delay. \paragraph{Asynchronous action-chunk execution.} Runtime methods such as Temporal Ensembling, BID, RTC, and A2C2 improve chunk transitions or correct actions during deployment \citep{zhao2023learning,liu2025bidirectional,black2026real,sendai2025leave}. Training-time methods including VLASH, TT-RTC, AsyncVLA, and REMAC adapt the policy to delayed execution through future-state conditioning, action-prefix conditioning, selective refinement, or masked chunk correction \citep{tang2025vlash,black2025trainingtimeactionconditioningefficient,jiang2511asyncvla,wang2026real}. DEFLECT differs by explicitly contrasting action candidates generated from
future and stale observations under the same deployable input. This supervision is constructed offline and leaves the asynchronous runtime unchanged.
\section{Implementation Details}
\label{app:implementation}

\paragraph{Preference-pair generation.}
Both the earlier input $(s_t,o_t)$ and the later input $(s_{t+\delta},o_{t+\delta})$ are taken from the same recorded expert trajectory. A frozen copy $\pi_{\mathrm{ref},b}$ of the SFT policy $b$ generates $A^{+}$ from the later input and $A^{-}$ from the earlier input using identical Gaussian noise and the same ODE schedule. These chunks are model predictions, not the expert actions recorded in the dataset. The expert actions are retained for the SFT anchor (Algorithm~\ref{alg:deflect_step}). We set the maximum temporal contrast horizon to $d_{\max}{=}4$ on both benchmarks and sample $\delta \sim \mathrm{Unif}\{1,\ldots,d_{\max}\}$; $\delta{=}0$ is excluded because $A^{+}$ and $A^{-}$ would coincide and degenerate the pair. The original supervised sampler $\mathcal{S}_b$ provides the deployment input and SFT target of each policy (Eq.~\eqref{eq:base_sft_sample}). For standard-SFT SmolVLA and GR00T N1.7, it uses the current input and the current expert action chunk. For the \textsc{VLASH}-based Kinetix, \pioh{}, and real-robot settings, it follows the original delay-aware context and target construction, including a context delay sampled from $\{0,\ldots,4\}$. The temporal contrast horizon $\delta$ is sampled independently of any context delay used by $\mathcal{S}_b$. Deployment delays are evaluated over $d \in \{0,\ldots,4\}$. For LIBERO, the frozen reference policy processes earlier and later inputs from recorded expert trajectories, and its outputs are cached as preference pairs before post-training. Training samples these cached pairs with replacement. With $2{,}000$ gradient steps, \pioh{} uses $16{,}000$ pairs per suite at batch size $8$, and GR00T N1.7 uses $8{,}000$ at batch size $4$, counting repeated uses. These counts describe training usage, not the number of distinct pairs in the cache. For Kinetix, preference chunks are generated as needed during training from inputs in recorded expert trajectories, rather than loaded from a precomputed cache.

\paragraph{Future SFT.}
For the GR00T N1.7 baseline, Future SFT uses the future-generated chunk $A_t^+$ as a supervised target under the deployment input $c_{t,b}^{\mathrm{dep}}$, applying the flow-matching loss of Eq.~\eqref{eq:fm_loss} to this target. The stale-generated chunk $A_t^-$ and the preference loss of Eq.~\eqref{eq:dpo_fm} are not used. We keep the original expert-target SFT term alongside the loss on the future-generated targets.

\paragraph{Training.}
We use AdamW with a $1{,}000$-step linear warmup followed by cosine decay to $1\%$ of the peak learning rate. VLA post-training on LIBERO and the real-robot tasks runs for $2{,}000$ gradient steps. Kinetix uses the $24$-epoch schedule. We set $\beta{=}1.0$ and $\lambda_{\mathrm{SFT}}{=}1.0$. Model-specific settings are listed in Table~\ref{tab:hyperparams}. The DPO weight $\lambda_{\mathrm{DPO}}$ is tuned per benchmark. On Kinetix we sweep $\lambda_{\mathrm{DPO}} \in \{0, 0.01, 0.02, 0.05, 0.1, 0.2, 0.5\}$ and select $0.02$ from the robust band $\lambda \in [0.01, 0.05]$. On LIBERO we use $\lambda_{\mathrm{DPO}}{=}0.1$.

\paragraph{Flow-matching loss reduction.}
For each example, the flow-matching loss of Eq.~\eqref{eq:fm_loss} sums squared errors over the chunk-horizon and action-dimension axes. Using a per-element mean instead of a sum divides the preference margin by the number of included elements, changing the effect of $\beta$ unless it is adjusted accordingly. For example, the divisors are $16$ for the Kinetix configuration with shape $8 \times 2$ and $350$ for the \pioh{} LIBERO configuration with shape $50 \times 7$. One flow time $\tau$ and one Gaussian noise sample $\epsilon$ are drawn per preference pair and shared across both candidates, the trainable and reference policies, and the SFT anchor. The outer DPO loss in Eq.~\eqref{eq:dpo_fm} is averaged over the minibatch.

\begin{table}[h]
\centering
\small
\caption{\textbf{Post-training settings for each benchmark and backbone.} The warmup lasts $1{,}000$ gradient steps in all settings. VLA post-training runs for $2{,}000$ gradient steps, while Kinetix uses $24$ epochs. Real-robot data collection is described in Appendix~\ref{app:realrobot}.}
\label{tab:hyperparams}
\setlength{\tabcolsep}{4pt}
\resizebox{\linewidth}{!}{%
\begin{tabular}{lccccc}
\toprule
 & Kinetix & \multicolumn{3}{c}{LIBERO} & Real robot \\
\cmidrule(lr){3-5}
Backbone
    & Flow policy
    & \pioh{}
    & \shortstack{GR00T\\N1.7}
    & SmolVLA
    & \pioh{} \\
\midrule
SFT initialization
    & \textsc{VLASH}
    & \textsc{VLASH}
    & Standard
    & Standard
    & \textsc{VLASH} \\
Trainable parameters
    & $3$M
    & $693$M
    & $1620$M
    & $97$M
    & $693$M \\
Training schedule
    & $24$ epochs
    & $2{,}000$ steps
    & $2{,}000$ steps
    & $2{,}000$ steps
    & $2{,}000$ steps \\
Batch size
    & $512$
    & $8$
    & $4$
    & $8$
    & $8$ \\
Peak learning rate
    & $3 \times 10^{-4}$
    & $1 \times 10^{-6}$
    & $1 \times 10^{-5}$
    & $1 \times 10^{-6}$
    & $1 \times 10^{-5}$ \\
Chunk length $H$
    & $8$
    & $50$
    & $16$
    & $50$
    & $50$ \\
DPO weight $\lambda_{\mathrm{DPO}}$
    & $0.02$
    & $0.1$
    & $0.1$
    & $0.1$
    & $0.02$ \\
\bottomrule
\end{tabular}%
}
\end{table}

All main-paper numbers, figures, and ablations on Kinetix use the $24$-epoch cosine setting. The VLM backbone is frozen for LIBERO, while Kinetix uses full fine-tuning.

\paragraph{Evaluation.}
On LIBERO, each method is evaluated with $500$ episodes per suite and delay for all three backbones, using MuJoCo 3.3.7. On Kinetix, each method is evaluated with $1{,}024$ rollouts per task for each delay and execution-horizon setting. Kinetix results are averaged over all $12$ tasks.

\paragraph{Benchmark protocols.}
We use the standard reward, termination, episode-length, success criterion, and action-space conventions from Kinetix~\citep{matthews2410kinetix} and LIBERO~\citep{liu2023libero} without modification. No task curriculum, action rescaling, or reward shaping is introduced for either benchmark.

\paragraph{Baselines.}
Main-table baselines (Naive, \textsc{VLASH}, \textsc{RTC}, \textsc{BID}) are all evaluated on the released \textsc{VLASH} checkpoint. \textsc{RTC} and \textsc{BID} are test-time chunk-level schedulers that replace the naive asynchronous rollover. \textsc{RTC} inpaints the residual chunk prefix at inference boundaries, and \textsc{BID} resamples chunks under forward-contrast and backward-coherence criteria. Both rely on the residual chunk prefix from the previous prediction, which shrinks to 1--3 steps at $d \geq 5$ and becomes too short to support either method. Other concurrent VLA work includes architectural variants (HybridVLA~\citep{liu2026hybridvla}, WorldVLA~\citep{cen2506worldvla}, OpenVLA-OFT~\citep{kim2016fine}, TinyVLA~\citep{wen2025tinyvla}, $\pi_0$-FAST~\citep{pertsch2025fast}, ManiFlow~\citep{yan2025maniflow}) and inference-acceleration methods (DeeR-VLA~\citep{yue2024deer}, SnapFlow~\citep{luan2026snapflow}, MP1~\citep{sheng2026mp1}). These target architectural design or inference cost rather than delay-robust execution and are orthogonal to our refinement.
\section{Algorithm: One Gradient Step of \textsc{DEFLECT}}
\label{app:algorithm}

Algorithm~\ref{alg:deflect_step} spells out one gradient step of \textsc{DEFLECT}: pair construction, shared $(\tau,\epsilon)$ sampling, the FM-loss evaluation, and the joint SFT+DPO update. In the LIBERO implementation, preference pairs are precomputed from recorded expert trajectories and loaded from the cache during post-training, rather than regenerated at every gradient step.

\begin{algorithm}[t] \caption{One gradient step of \textsc{DEFLECT}.} \label{alg:deflect_step} \begin{algorithmic}[1] \REQUIRE SFT-trained policy $b$; its original supervised sampler $\mathcal{S}_b$; frozen reference copy $\pi_{\mathrm{ref},b}$; trainable parameters $\theta$ initialized from $b$; demonstration dataset $\mathcal{D}$; maximum temporal contrast horizon $d_{\max}$; loss weights $\lambda_{\mathrm{SFT}},\lambda_{\mathrm{DPO}},\beta$. \STATE Sample timestep $t$ and retrieve instruction $l_t$. \STATE Draw the original base-policy training sample $(c^{\mathrm{dep}}_{t,b},A^{\mathrm{exp}}_{t,b}) \sim\mathcal{S}_b(\mathcal{D},t)$. \STATE Sample temporal contrast horizon $\delta\sim\mathrm{Unif}\{1,\ldots,d_{\max}\}$. \COMMENT{$\delta=0$ gives $A^+=A^-$} \STATE Sample shared reference-policy randomness $\xi$. \STATE Generate $A^+ \leftarrow \pi_{\mathrm{ref},b} ((s_{t+\delta},o_{t+\delta}),l_t;\xi)$ and $A^- \leftarrow \pi_{\mathrm{ref},b} ((s_t,o_t),l_t;\xi)$. \STATE Sample $\tau\sim\mathrm{Unif}[0,1]$ and Gaussian $\epsilon$, shared across the preference pair and SFT anchor, and construct $\mathbf{x}_\tau^\pm =\tau A^\pm+(1-\tau)\epsilon$ and $\mathbf{x}_\tau^{\mathrm{exp}} =\tau A^{\mathrm{exp}}_{t,b}+(1-\tau)\epsilon$. \STATE Query $v_\theta^\pm =v_\theta(\mathbf{x}_\tau^\pm,c^{\mathrm{dep}}_{t,b},\tau)$ and the corresponding reference velocities $v_{\mathrm{ref}}^\pm$. \STATE Compute $r_\theta^\pm =-\|v_\theta^\pm-(A^\pm-\epsilon)\|_2^2$ and $r_{\mathrm{ref}}^\pm =-\|v_{\mathrm{ref}}^\pm-(A^\pm-\epsilon)\|_2^2$. \STATE $\mathcal{L}_{\mathrm{DPO}} \leftarrow -\log\sigma\!\left( \beta\left[ (r_\theta^+-r_\theta^-) - (r_{\mathrm{ref}}^+-r_{\mathrm{ref}}^-) \right]\right)$. \STATE $\mathcal{L}_{\mathrm{FM}} \leftarrow \|v_\theta( \mathbf{x}_\tau^{\mathrm{exp}}, c^{\mathrm{dep}}_{t,b}, \tau) - (A^{\mathrm{exp}}_{t,b}-\epsilon)\|_2^2$. \STATE Update $\theta$ using $\nabla_\theta[ \lambda_{\mathrm{SFT}}\mathcal{L}_{\mathrm{FM}} + \lambda_{\mathrm{DPO}}\mathcal{L}_{\mathrm{DPO}} ]$. \end{algorithmic} \end{algorithm}

\section{Flow-Matching Loss as a Likelihood Surrogate for DPO}
\label{app:fm_surrogate}

Section~\ref{sec:deployment_dpo} uses the flow-matching objective as a
tractable surrogate for the action log likelihood required by DPO. We clarify
here the distinction between the expected flow-matching objective, its
variational interpretation, and the Monte Carlo estimator used in training.

\paragraph{Expected flow-matching objective.}
For an action chunk $A$, conditioning input $c$, flow time
$\tau\sim\mathrm{Unif}[0,1]$, and Gaussian noise
$\epsilon\sim\mathcal{N}(0,I)$, define
\begin{equation}
\begin{aligned}
    \ell_{\mathrm{FM}}
    (\theta;A,c,\tau,\epsilon)
    &=
    \left\|
        v_\theta(
        \tau A+(1-\tau)\epsilon,
        \tau,c)
        -(A-\epsilon)
    \right\|_2^2,\\
    \overline{\mathcal L}_{\mathrm{FM}}(\theta;A,c)
    &=
    \mathbb E_{\tau,\epsilon}
    \left[
        \ell_{\mathrm{FM}}
        (\theta;A,c,\tau,\epsilon)
    \right].
\end{aligned}
\label{eq:expected_fm}
\end{equation}
Under the Gaussian-source, linear-interpolant formulation, prior work
motivates this objective through a variational lower bound on the conditional
log likelihood: Diffusion-DPO substitutes the diffusion ELBO for the log
likelihood~\citep{wallace2024diffusion}, and \citet{mcallister2026flow}
derive an analogous bound for this flow family while retaining a
$\theta$-dependent gap. We therefore use
$-\overline{\mathcal L}_{\mathrm{FM}}$ as an implicit likelihood score.
We do not identify it with the exact value of
$\log\pi_\theta(A\mid c)$.

\paragraph{Reference-calibrated preference surrogate.}
Define
\begin{equation}
    r^{\mathrm{FM}}_\theta(A\mid c)
    =
    -\overline{\mathcal L}_{\mathrm{FM}}(\theta;A,c).
\end{equation}
The corresponding pairwise margin is
\begin{equation}
\begin{aligned}
    M_{\mathrm{FM}}
    ={}&
    \left[
        r^{\mathrm{FM}}_\theta(A^+\mid c)
        -
        r^{\mathrm{FM}}_\theta(A^-\mid c)
    \right]\\
    &-
    \left[
        r^{\mathrm{FM}}_{\mathrm{ref}}(A^+\mid c)
        -
        r^{\mathrm{FM}}_{\mathrm{ref}}(A^-\mid c)
    \right].
\end{aligned}
\label{eq:fm_margin_surrogate}
\end{equation}
This has the same reference-calibrated pairwise structure as the standard
DPO log-ratio margin, but should generally be interpreted as its
flow-matching surrogate. Under the stronger assumption that the expected
flow-matching objective differs from the exact negative log likelihood only
by the same model-independent offset for the trainable and reference
policies, these offsets cancel and
$M_{\mathrm{FM}}$ coincides with the DPO log-ratio margin. Our optimization
does not require this stronger equality.

\paragraph{Monte Carlo realization.}
During training, we estimate
$\overline{\mathcal L}_{\mathrm{FM}}$ using one sampled pair
$(\tau,\epsilon)$:
\begin{equation}
    \widehat{\mathcal L}_{\mathrm{FM}}
    (\theta;A,c)
    =
    \ell_{\mathrm{FM}}
    (\theta;A,c,\tau,\epsilon),
    \qquad
    \mathbb E[
        \widehat{\mathcal L}_{\mathrm{FM}}
    ]
    =
    \overline{\mathcal L}_{\mathrm{FM}}.
\end{equation}
The same $(\tau,\epsilon)$ is shared across $A^+$ and $A^-$ and across the
trainable and reference policies. This common-random-number construction
reduces the variance of the pairwise difference, but it does not remove
Monte Carlo error or turn the estimator into an exact likelihood ratio.
Since $-\log\sigma(z)$ is convex in $z$, by Jensen's inequality the
expected single-sample training loss upper-bounds the loss evaluated at the
expected margin, the same estimator structure used in
Diffusion-DPO~\citep{wallace2024diffusion}.

\paragraph{Scope.}
Accordingly, DEFLECT optimizes a reference-calibrated flow-matching
surrogate motivated by the variational likelihood interpretation. We do not
claim exact recovery of the action log likelihood or of the original DPO
log-ratio objective.

\section{Generalization across Contrast Horizons}
\label{app:zeroshot}

We compare three temporal contrast-horizon settings. \textbf{Model A} uses $\delta \in \{1,2\}$, while the main \textbf{Model B} uses $\delta \in \{1,\ldots,4\}$. \textbf{Model C} uses $\delta \in \{1,\ldots,7\}$ and also changes the asynchronous interval to $8$. At deployment delay $d{=}4$, Model A improves success over \textsc{VLASH} by $3.2$ percentage points despite using contrast horizons of at most two steps during training. This supports generalization beyond the training contrast horizons. Model C achieves $88.2\%$ average success, compared with $89.2\%$ for Model B. Because Model C changes both the contrast-horizon range and the asynchronous interval, this difference cannot be attributed to the contrast-horizon range alone.

\begin{figure}[h]
\centering
\includegraphics[width=0.75\linewidth]{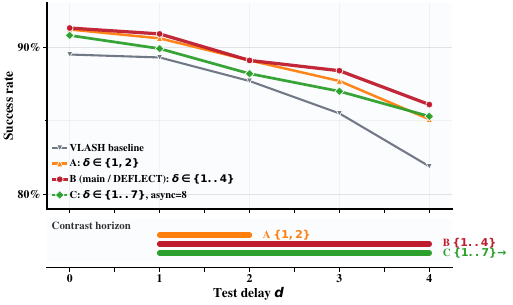}
\caption{\textbf{Generalization beyond the training contrast horizons.} Model A beats \textsc{VLASH} at every deployment delay despite never using a contrast horizon $\delta \geq 3$ in its pairs. The strip below marks each variant's contrast-horizon range; Model C's extends past the plotted range (arrow).}
\label{fig:zeroshot}
\end{figure}

\begin{table}[h]
\centering
\small
\caption{\textbf{Contrast-horizon generalization} on Kinetix (success rate $\%$). Column subscripts give each variant's $\delta$-range; rows are deployment delays. Model C also changes the asynchronous interval to $8$.}
\label{tab:zeroshot}
\begin{tabular}{ccccc}
\toprule
$d$ & \textsc{VLASH} & A$_{\{1,2\}}$ & B$_{\{1,\dots,4\}}$ & C$_{\{1,\dots,7\}}$ \\
\midrule
0 & $89.5$ & $91.2$ & $91.3$ & $90.8$ \\
1 & $89.3$ & $90.6$ & $90.9$ & $89.9$ \\
2 & $87.7$ & $89.1$ & $89.1$ & $88.2$ \\
3 & $85.5$ & $87.7$ & $88.4$ & $87.0$ \\
4 & $81.9$ & $85.1$ & $86.1$ & $85.3$ \\
\midrule
avg & $86.8$ & $88.7$ & $\mathbf{89.2}$ & $88.2$ \\
\bottomrule
\end{tabular}
\end{table}

\section{Design Ablations}
\label{app:ablations}

Two design choices define \textsc{DEFLECT}: the SFT anchor is kept alongside the preference term, and both candidates are scored under the same deployment input rather than each under its own generation context. Table~\ref{tab:design_ablations} isolates both on Kinetix; every variant starts from the same \textsc{VLASH} checkpoint and shares the training schedule of the main run.

Removing the SFT anchor reduces average success to $15.1\%$ and success at $d{=}4$ to $10.6\%$, showing that the preference objective alone is insufficient in this setting. Scoring each candidate under its own generation input ($A^{+}$ under the future observation, $A^{-}$ under the stale one) instead of the shared deployment input costs $-0.7$ on average and $-1.3$ at $d{=}4$, and the gap widens monotonically with delay ($-0.3$, $-0.5$, $-0.7$, $-0.9$, $-1.3$ over $d{=}0,\ldots,4$), consistent with deployment-input alignment mattering more as the two generation contexts diverge.

\begin{table}[h]
\centering
\small
\caption{\textbf{Design ablations} on Kinetix (success rate $\%$).}
\label{tab:design_ablations}
\begin{tabular}{lcc}
\toprule
Variant & avg($d{=}0\text{-}4$) & $d{=}4$ \\
\midrule
\textsc{VLASH} base                    & $86.8$ & $81.9$ \\
\textsc{DEFLECT} w/o SFT anchor        & $15.1$ & $10.6$ \\
\textsc{DEFLECT}, scoring under each generation input & $88.4$ & $84.8$ \\
\textbf{\textsc{DEFLECT} (full)}       & $\mathbf{89.2}$ & $\mathbf{86.1}$ \\
\bottomrule
\end{tabular}
\end{table}

\section{Effect of Temporal Preference-Label Direction}
\label{app:label_direction}

\paragraph{Experimental question.}
We examine whether DEFLECT's training gains depend on the temporal ordering
of preference labels. We compare the original DEFLECT ordering
$A^+ \succ A^-$ with randomized and fully reversed orderings, as well as
Continued SFT, which retains the supervised anchor without the preference term.
The original, randomized, and reversed variants test the role of label direction;
Continued SFT provides the comparison against supervised continuation alone.

\paragraph{Controlled training setup.}
The experiment uses the implementation of DEFLECT described in this paper.
Within each suite, we hold the initialization, frozen reference model,
supervised-anchor construction, training budget, precision, and hyperparameters
fixed. The three preference variants use the same candidate-pair dataset and
change only the label ordering: the original ordering prefers $A^+$ to $A^-$,
random labels randomize this ordering, and reversed labels prefer $A^-$ to
$A^+$ for every pair. Swapping labels affects only the preference term and
leaves the SFT anchor's supervised targets unchanged. Continued SFT sets
$\lambda_{\mathrm{DPO}}=0$.

\paragraph{Evaluation.}
We evaluate GR00T N1.7 on LIBERO-Object, LIBERO-Goal, LIBERO-Spatial, and
LIBERO-10 at inference delays $d\in\{1,2,3,4\}$. Each delay column in
Table~\ref{tab:label_direction_single_seed} is the equally weighted mean success
rate over the four suites, and Avg. additionally averages equally over the four
delays. Differences are computed before rounding.

\begin{table}[t]
  \centering
  \small
  \caption{\textbf{Effect of temporal preference-label direction on GR00T N1.7.}
  Success rates (\%) are averaged equally over the four LIBERO suites at each
  inference delay. Avg. additionally averages equally over $d=1,2,3,4$.
  The three preference variants differ only in label ordering, and Continued SFT
  disables the preference term. Original-label DEFLECT and Continued SFT
  results are reused from Table~\ref{tab:libero_main}(a); randomized and
  reversed labels are additional controls using the same training and
  evaluation protocol.}
  \label{tab:label_direction_single_seed}
  \begin{tabular}{lrrrrr}
    \toprule
    Method & $d=1$ & $d=2$ & $d=3$ & $d=4$ & Avg. \\
    \midrule
    DEFLECT (original labels) & \textbf{93.60} & \textbf{91.05} & \textbf{85.30} & \textbf{73.10} & \textbf{85.76} \\
    Continued SFT             & 91.65 & 85.40 & 75.55 & 62.25 & 78.71 \\
    Random labels             & 92.50 & 83.75 & 71.10 & 53.20 & 75.14 \\
    Reversed labels           & 84.95 & 71.95 & 55.10 & 36.50 & 62.13 \\
    \bottomrule
  \end{tabular}
\end{table}

\paragraph{Results.}
The original temporal labels achieve 85.76\% average success, exceeding
Continued SFT by 7.05 percentage points, randomized labels by 10.63 points, and
reversed labels by 23.64 points. The margin over randomized labels increases
from 1.10 points at $d=1$ to 19.90 points at $d=4$; the margin over Continued
SFT increases from 1.95 to 10.85 points over the same delays. Averaged over
the four suites and delays, the ordering is
DEFLECT $>$ Continued SFT $>$ random labels $>$ reversed labels.

\paragraph{Interpretation.}
The label-direction ablation demonstrates that temporal ordering is a
consequential component of DEFLECT: preserving the original ordering improves
performance over both supervised continuation and randomized or reversed
preferences, with larger gains at longer delays.

\section{Real-Robot Per-Sub-Goal Breakdown}
\label{app:realrobot}

\paragraph{Hardware setup.}
We use a bimanual AgileX dual-arm platform with $6$ DoF per arm, three Intel RealSense D435i cameras, one head-mounted overhead and one wrist-mounted on each arm, $30$\,Hz control, and a joint-position action space of $7$ dimensions per arm and $14$ dimensions bimanually, namely $6$ joint targets plus $1$ gripper open/close per arm. The reference policy is a $\pi_{0.5}$ checkpoint we trained in-house on this platform following the \vlash{} training procedure on our own teleoperation demonstrations; it is not the released LIBERO checkpoint used in the simulation experiments. Policy inference runs on a single NVIDIA RTX $5090$. No additional delay is injected in the real-robot experiments. On the RTX 5090, model inference takes $57.1$\,ms, which at $30$\,Hz rounds up to $2$ control steps (Table~\ref{tab:latency}). All three methods use the same model architecture, number of denoising steps, and asynchronous controller. We compare them under this deployment setup rather than sweeping over injected delays. One implementation detail is worth recording for anyone reproducing these timings: openpi pads absent camera slots with black images and runs the vision encoder over all three regardless, so reducing the camera count does not reduce latency in this stack ($61.2$\,ms with one camera versus $61.1$\,ms with three, the difference being only the two skipped resizes).

\paragraph{Demonstration data.}
All demonstrations are collected by leader--follower teleoperation on the same platform at $30$\,Hz, recording the three camera streams at $480 \times 640$ together with the $14$-dimensional joint-position state and action. Table~\ref{tab:realrobot_data} lists the per-task budget. No demonstrations are shared across tasks, and no simulation data is used.

\begin{table}[h]
\centering
\small
\caption{\textbf{Real-robot demonstration data.} Each task is trained from its own demonstrations only.}
\label{tab:realrobot_data}
\begin{tabular}{lccc}
\toprule
Task & Demonstrations & Total duration & Mean episode \\
\midrule
\emph{Conveyor-I}     & $199$ & $75.0$\,min  & $22.6$\,s \\
\emph{Conveyor-II}    & $149$ & $55.7$\,min  & $22.4$\,s \\
\emph{Whack-a-Mole}   & $192$ & $122.7$\,min & $38.3$\,s \\
\bottomrule
\end{tabular}
\end{table}

\paragraph{Training setup.}
For each task we first train the $\pi_{0.5}$ reference policy on that task's demonstrations, then apply \textsc{DEFLECT} to it; the \pioh{} and \vlash{} rows of Tables~\ref{tab:realrobot_conveyor_simple}--\ref{tab:realrobot_conveyor} are the corresponding checkpoints of the same pipeline, so all three methods share one dataset, one backbone, and one control stack. The policy predicts chunks of $H{=}50$ actions and the flow-matching ODE is integrated with $10$ denoising steps, both at training and at deployment. The \textsc{DEFLECT} stage runs for $2{,}000$ gradient steps at batch size $8$ and learning rate $1 \times 10^{-5}$, with $\beta{=}1.0$, $\lambda_{\mathrm{SFT}}{=}1.0$, and $\lambda_{\mathrm{DPO}}{=}0.02$, updating only the action expert while the VLM backbone stays frozen. The learning rate uses a $1{,}000$-step linear warmup followed by cosine decay to $1\%$ of its peak value. The contrast horizon is drawn from $\delta \in \{1,\ldots,4\}$, the same range used in simulation, and pairs are generated offline from the recorded demonstrations without any additional robot time.

\paragraph{Conveyor-I (simple, two sub-goals).}
The simpler variant scores two sequential sub-goals: S1 (left arm places cup on conveyor) and S2 (right arm retrieves cup from conveyor). Full-task success is the conjunction.

\begin{table}[h]
\centering
\small
\caption{\textbf{Conveyor-I per-sub-goal success} ($N{=}30$ trials per method). All methods complete S1 (cup placement) without failure, so the only failure mode on this task is S2 (cup retrieval). \textsc{DEFLECT} improves S2 success by $+10$ pp over \vlash{} ($96.7$ vs.\ $86.7$) and by $+20$ pp over \pioh{} ($96.7$ vs.\ $76.7$).}
\label{tab:realrobot_conveyor_simple}
\begin{tabular}{lccc}
\toprule
Method & S1: cup on conveyor & S2: cup off conveyor & Full task ($\wedge$ S1, S2) \\
\midrule
\pioh{}                          & $30/30$ ($100.0$) & $23/30$ ($76.7$) & $23/30$ ($76.7$) \\
\vlash{}                         & $30/30$ ($100.0$) & $26/30$ ($86.7$) & $26/30$ ($86.7$) \\
\textbf{\textsc{DEFLECT} (ours)} & $\mathbf{30/30}$ ($\mathbf{100.0}$) & $\mathbf{29/30}$ ($\mathbf{96.7}$) & $\mathbf{29/30}$ ($\mathbf{96.7}$) \\
\bottomrule
\end{tabular}
\end{table}

\paragraph{Conveyor-II (full, three sub-goals).}
The harder variant inserts a middle sub-goal (S2: select the red rather than green cube and place into cup), so the scoring becomes S1 (cup on conveyor), S2 (correct cube into cup), S3 (cup off conveyor).
Sub-goal outcomes are recorded separately over all $30$ trials.
Later sub-goals are still evaluated when an earlier sub-goal fails.
Full-task success requires all three sub-goals to succeed in the same trial.

\begin{table}[h]
\centering
\small
\caption{\textbf{Conveyor-II per-sub-goal success} ($N{=}30$ trials per method, scored on three sequential sub-goals: \textbf{S1} cup on conveyor, \textbf{S2} correct cube into cup, \textbf{S3} cup off conveyor; full task is the conjunction). The \pioh{} failures are concentrated on S3 (cup retrieval), with a smaller drop on S2 (cube selection). \vlash{} closes both gaps but still misses $5/30$ on S3 and $1/30$ on S2.}
\label{tab:realrobot_conveyor}
\begin{tabular}{lcccc}
\toprule
Method & S1 & S2 & S3 & Full task \\
\midrule
\pioh{}                 & $30/30$ ($100.0$) & $22/30$ ($73.3$) & $15/30$ ($50.0$) & $14/30$ ($46.7$) \\
\vlash{}                & $30/30$ ($100.0$) & $29/30$ ($96.7$) & $25/30$ ($83.3$) & $25/30$ ($83.3$) \\
\textbf{\textsc{DEFLECT} (ours)} & $\mathbf{30/30}$ ($\mathbf{100.0}$) & $\mathbf{30/30}$ ($\mathbf{100.0}$) & $\mathbf{27/30}$ ($\mathbf{90.0}$) & $\mathbf{27/30}$ ($\mathbf{90.0}$) \\
\bottomrule
\end{tabular}
\end{table}

\section{Training-Time Cost of \textsc{DEFLECT}}
\label{app:runtime}

\textsc{DEFLECT} leaves inference unchanged. The deployed policy retains the flow-matching ODE, deployment input, and step budget of its SFT initialization, while the additional cost is confined to offline training (Table~\ref{tab:train_cost}). Counts are per gradient step and per network call, with the batch composition stated here rather than implied. On Kinetix, preference pairs are sampled on the fly, so each gradient step pays $11$ frozen-reference forwards ($2$ chunk generations at $5$ denoising steps each, plus one reference-velocity evaluation for the margin in Eq.~\eqref{eq:dpo_fm}).
On LIBERO, preference pairs are generated by the frozen reference policy from recorded expert trajectories and cached before post-training. The two chunks in each preference pair share the deployment context, flow time, and Gaussian noise, but each is evaluated at its own interpolated input $A_\tau^{\pm}$ of Eq.~\eqref{eq:fm_loss}. A batch of four preference pairs therefore requires velocity predictions for eight candidate inputs. These eight inputs are concatenated into one network call, so each gradient step makes two calls that carry gradients, one over the eight candidate inputs and one for the SFT anchor term, plus one gradient-free reference call over the same eight inputs at the same $(\tau,\epsilon)$.
For GR00T N1.7 the cache holds $12{,}000$ pairs per suite over four suites, and the released checkpoint integrates the flow ODE in $4$ steps. Generating both chunks for all $48{,}000$ cached pairs with four flow steps requires $48{,}000 \times 2 \times 4 = 384{,}000$ single-candidate forward evaluations. Batched generation can process multiple such evaluations in one network call. All reference passes carry no gradient.

\begin{table}[h]
\centering
\small
\caption{\textbf{Training and inference cost} relative to continued SFT on the same backbone (fwd $=$ forward pass). Per-step counts refer to network calls with the batch composition stated in the text; the one-time cost is reported in single-candidate forward evaluations. The LIBERO row is stated for GR00T N1.7, and its one-time column covers all four suites.}
\label{tab:train_cost}
\begin{tabular}{lccc}
\toprule
 & SFT & \textsc{DEFLECT} & one-time \\
\midrule
Kinetix, per gradient step & $1$ fwd$+$bwd & $1$ fwd$+$bwd $\,+\,11$ ref fwd & --- \\
LIBERO, per gradient step  & $1$ fwd$+$bwd & $2$ fwd$+$bwd $\,+\,1$ ref fwd  & $\approx 3.8 \times 10^{5}$ ref fwd \\
Inference                  & $1\times$     & $1\times$                       & --- \\
\bottomrule
\end{tabular}
\end{table}

\end{document}